\documentclass{article}
\usepackage{spconf,amsmath,graphicx,hyperref}
\usepackage{multirow}
\usepackage{booktabs}
\usepackage{amssymb} 

\title{VocalNet-M2: Advancing Low-Latency Spoken Language Modeling via Integrated Multi-Codebook Tokenization and Multi-Token Prediction}
\name{Yuhao Wang$^{1,2}$, Ziyang Cheng$^{1}$, Heyang Liu$^{1,2}$, Ronghua Wu$^2$, Qunshan Gu$^{2}$, Yanfeng Wang$^1$, Yu Wang$^{1\dagger}$\thanks{$^\dagger$Corresponding author}}
\address{$^1$Shanghai Jiao Tong University \\
$^2$Ant Group}

\begin{document}
\ninept
\maketitle
\begin{abstract}
Current end-to-end spoken language models (SLMs) have made notable progress, yet they still encounter considerable response latency. This delay primarily arises from the autoregressive generation of speech tokens and the reliance on complex flow-matching models for speech synthesis. To overcome this, we introduce VocalNet-M2, a novel low-latency SLM that integrates a multi-codebook tokenizer and a multi-token prediction (MTP) strategy. Our model directly generates multi-codebook speech tokens, thus eliminating the need for a latency-inducing flow-matching model. Furthermore, our MTP strategy enhances generation efficiency and improves overall performance. Extensive experiments demonstrate that VocalNet-M2 achieves a substantial reduction in first chunk latency (from approximately 725ms to 350ms) while maintaining competitive performance across mainstream SLMs. This work also provides a comprehensive comparison of single-codebook and multi-codebook strategies, offering valuable insights for developing efficient and high-performance SLMs for real-time interactive applications.
\end{abstract}

\section{Introduction}
\label{sec:intro}

In recent years, end-to-end spoken language models (SLMs) have made rapid progress, marking a significant milestone in generative AI~\cite{xu2025qwen2,wu2025step,li2025baichuan,zeng2024glm}. These models learn to directly model discrete speech tokens derived from a speech tokenizer, endowing Large Language Models (LLMs) with the dual ability to comprehend and generate both text and speech, which is particularly valuable for applications requiring natural and fluent human-computer dialogue, such as voice assistants and real-time conversational agents.

Current approaches for multimodal modeling of text and speech in SLMs can be broadly categorized into two paradigms. The first is the speech-native multimodal approach, which directly extends an LLM's vocabulary to include tokens from speech corpora~\cite{wu2025step,li2025baichuan,zeng2024glm,chen2024slam,xie2024mini}. The second is the modality-alignment approach~\cite{yu2025salmonn, chen2025minmo,wang2025vocalnet}. This method leverages pre-trained LLMs and integrates separate speech input/output modules through efficient cross-modal alignment. This reduces the need for extensive training from scratch, building upon existing LLM capabilities.

Despite these advancements, a significant challenge for current open-source SLMs is high response latency, which severely degrades user experience in speech interactions. This latency primarily arises from two sources: the autoregressive generation of both text and speech tokens by the Transformer decoder, and the subsequent conversion of speech tokens into waveforms. Most SLMs are trained to generate a single codebook of semantic speech tokens~\cite{chen2025minmo,wang2025vocalnet,zeng2024glm,chen2024slam,fang2025llama}. These tokens are then converted into a mel spectrogram by a flow-matching model, which a vocoder finally synthesizes into an audio waveform~\cite{du2024cosyvoice}. While semantic tokens effectively capture linguistic content, their limited acoustic information necessitates a flow-matching model for detailed acoustic reconstruction. Although this simplifies the speech modeling task for the LLM and can produce high-quality speech, it introduces a critical bottleneck. The heavy reliance on the flow-matching model for acoustic reconstruction leads to substantial computational overhead and, critically, significant inference latency. This poses a major challenge for real-time interactive applications where low-latency turn-taking is essential.

To mitigate this, a direct approach is to empower SLMs to generate multi-codebook speech tokens. These tokens inherently contain richer acoustic information, which could eliminate the need for a separate, latency-inducing flow-matching model. While Moshi~\cite{defossez2024moshi} has explored multi-codebook strategies, its application has been confined to the speech-native multimodal paradigm. Furthermore, a systematic analysis of the architectural design and a comprehensive comparison between single-codebook and multi-codebook strategies remain largely unexplored.

Therefore, in this work, we propose VocalNet-M2, a novel low-latency modality-alignment SLM incorporating a multi-codebook tokenizer and a multi-token prediction (MTP) strategy~\cite{gloeckle2024better}. Our main contributions are threefold:
\begin{itemize}
\item We propose a novel modality-aligned spoken language model architecture capable of directly generating \textbf{multi-codebook speech tokens}, eliminating the need for a flow-matching model and enabling more streamlined and efficient speech response generation.
\item We design a specialized \textbf{multi-token prediction strategy} tailored for multi-codebook generation, which significantly improves the performance of VocalNet-M2 while further reducing inference latency.
\item We conduct extensive experiments comparing single-codebook and multi-codebook strategies, revealing their respective strengths and trade-offs. Our findings provide actionable insights for future research on efficient and high-performance spoken language models.
\end{itemize}

\begin{figure*}[ht]
\begin{minipage}[b]{1.0\linewidth}
  \centering
  \centerline{\includegraphics[width=1.0\textwidth]{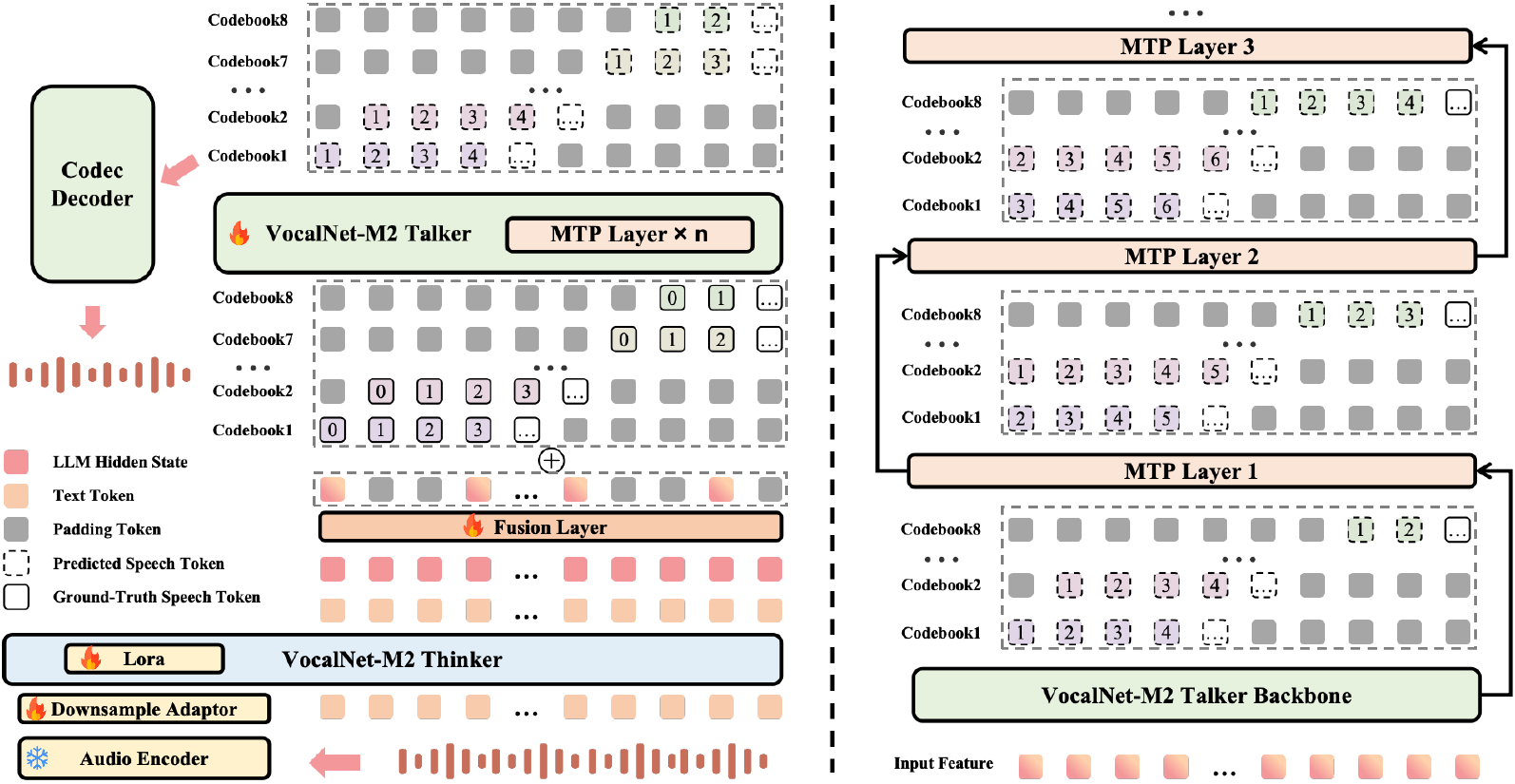}}
  \caption{Left: Overview of the VocalNet-M2 architecture. Right: The detailed architecture of VocalNet-M2 Talker.}
  \label{fig:vocalnet-m2}
\end{minipage}
\end{figure*}

\section{VocalNet-M2}
\label{sec:VocalNet-M2}

\subsection{Model Architecture}
VocalNet-M2 adopts a Thinker-Talker architecture~\cite{xu2025qwen2}, as illustrated in Figure~\ref{fig:vocalnet-m2} (Left).

Given a raw audio input \({x}^a\), it is first processed by an \textbf{Audio Encoder} and a \textbf{Downsample Adaptor}. This converts the raw audio into continuous representations \({r^{a}_{1:T}}\) that capture the high-quality features of the input. These representations are then fed into the \textbf{VocalNet-M2 Thinker}. The Thinker is responsible for autoregressively generating both the textual response tokens and their corresponding hidden states:
\begin{equation}
    ({t^{\text{text}}_{1:N}, h^{\text{text}}_{1:N}}) = \mathcal{T}_{\text{thinker}}({r^{a}_{1:T}})
\end{equation}
Here, \({t^{\text{text}}_{1:N}}\) represents the sequence of generated text tokens, and \({h^{\text{text}}_{1:N}}\) are their associated LLM hidden states, providing a rich semantic embedding for subsequent speech generation.

The \textbf{VocalNet-M2 Talker} is a multi-track autoregressive transformer decoder designed to generate speech tokens. VocalNet-M2 utilizes the XY-Tokenizer~\cite{gong2025xy} to extract eight codebook audio tokens. Consequently, as shown in Figure~\ref{fig:vocalnet-m2}, the Talker operates with eight distinct audio tracks (one for each codebook) and a semantic representation track as input. Its output consists of eight audio tracks, corresponding to the predicted tokens for each codebook.

Before being input to the Talker, the hidden states and text embeddings from the Thinker are processed by a \textbf{Fusion Layer}. This layer, composed of 2 linear layers, fuses the text embeddings and the Thinker's hidden states to create a unified semantic representation:
\begin{equation}
    {h^{\text{fused}}_{1:N}} = \mathrm{Linear}\Big{(}\sigma \big{(} \mathrm{Linear}(\mathrm{Emb}({t^{\text{text}}_{1:N}})|| {h^{\text{text}}_{1:N}})\big{)}\Big{)}
\end{equation}
Where $\mathrm{Emb}({t^{\text{text}}_{1:N}})$ denotes the embeddings of the text tokens, and $||$ signifies concatenation.

To address the inherent frequency mismatch between text and speech tokens, we first upsample the fused representation \({h^{\text{fused}}_{1:N}}\) to three times its original length. This upsampling aims to promote better temporal alignment between the semantic information and the audio tokens.  For the current decoding timestep \(t\), this upsampled sequence is then either truncated or padded with zeros to match the current audio tokens length \(t\), forming \({h^{\text{up}}_{1:t}}\):

\begin{equation}
{h^{\text{up}}_{1:t}} = \begin{cases}
[h^{\text{fused}}_{1}, \mathbf{0}, \mathbf{0}, \cdots, h^{\text{fused}}_{N}, \mathbf{0}, \mathbf{0}]_{1:t} & \text{if } 3N \geq t \\
[h^{\text{fused}}_{1}, \mathbf{0}, \mathbf{0}, \cdots, h^{\text{fused}}_{N}, \mathbf{0}, \mathbf{0}, \mathbf{0}, \cdots, \mathbf{0}]_{1:t} & \text{if } 3N < t
\end{cases}
\end{equation}

Then, the {VocalNet-M2 Talker} predicts the audio tokens at time \(t+1\). Its input consists of the upsampled hidden states (\({h^{\text{up}}_{1:t}}\)) and the embeddings of all previously predicted audio tokens (\(\sum_{j=1}^{8}\mathrm{Emb}(a^{\text{cb}j}_{1:t})\)). The Talker then outputs the eight audio tokens for time \(t+1\) via eight linear layers:

\begin{equation}
\{a^{\text{cb}j}_{t+1}\}_{j=1}^{8} = \mathcal{T}_{\text{talker}}(h^{\text{up}}_{1:t} + \sum_{j=1}^{8}\mathrm{Emb}(a^{\text{cb}j}_{1:t}))
\end{equation}

\vspace{-0.1cm}
This design allows VocalNet-M2 to generate audio tokens in a streaming manner by continuously feeding the \({h^{\text{up}}}\) sequence into the Talker. During training, the cross-entropy loss is computed for each codebook at each time step \(t\):
\begin{equation}
    \mathcal{L}_{talker} = -\sum_{t=0}^{M-1}\sum_{j=1}^{8}\mathrm{log}P(a^{\text{cb}j}_{t+1}|h^{up}_{1:t}, \{a^{\text{cb}i}_{1:t}\}_{i=1}^{8})
\end{equation}
Here, \(M\) is the length of audio token.

\begin{table*}[t]
\centering
\caption{Comparison between VocalNet-M2 and other mainstream SLMs. We evaluate their performance including text quality, speech quality, and first chunk latency. Note that Qwen2.5-Omni's first chunk latency was not measured due to the absence of officially provided streaming inference code. The latency results are shown in the formation `mean$\pm$standard error'.}
\resizebox{1\linewidth}{!}{
\begin{tabular}{lcccccccl}
\toprule
\multirow{2}{*}{\textbf{Model}} & \multicolumn{4}{c}{\textbf{text}} & \multicolumn{2}{c}{\textbf{speech}} & \multirow{2}{*}{\textbf{First chunk latency (ms)}} \\
\cmidrule(lr){2-5} \cmidrule(lr){6-7}
& \textbf{AlpacaEval} & \textbf{Llama Questions} & \textbf{TriviaQA} & \textbf{Web Questions} & \textbf{wer} & \textbf{utmos} \\
\hline
SLAM-Omni~\cite{chen2024slam} & 3.50 & 2.94 & 0.39 & 0.84 & 5.78 & 4.46 & 702.41 $\pm$ 30.30 \\
VocalNet-8B~\cite{wang2025vocalnet} & 7.12 & 7.95 & 6.24 & 6.48 & 3.64 & \textbf{4.49} & 556.00 $\pm$ 8.29 \\
GLM-4-Voice~\cite{zeng2024glm} & 5.86 & 7.74 & 4.95 & 5.56 & 11.90 & 4.23 & 1060.36 $\pm$ 2.36 \\
MiniCPM-o~\cite{MiniCPM-o-2.6} & 6.13 & 7.72 & \textbf{6.43} & \textbf{7.16} & 9.52 & 4.14 & 893.82 $\pm$ 81.80 \\
kimi-audio~\cite{ding2025kimi} & 6.49 & 8.10 & 6.15 & 7.10 & 14.71 & 2.87 & 1744.80 $\pm$ 139.99 \\
Qwen2.5-Omni~\cite{xu2025qwen2} & 6.01 & 7.90 & 5.89 & 6.88 & \textbf{2.31} & 4.34 & \textbackslash \\
\textbf{VocalNet-M2} & \textbf{7.29} & \textbf{8.33} & 6.13 & 6.65 & 6.07 & 4.31 & \textbf{348.86 $\pm$ 2.86} \\
\bottomrule
\end{tabular}
}
\label{tab:performance_metrics}
\end{table*}

\subsection{Multi-token Prediction (MTP)}
To enhance generation efficiency and capture local dependencies more effectively~\cite{wang2025vocalnet}, VocalNet-M2 incorporates a MTP mechanism. As shown in Figure~\ref{fig:vocalnet-m2} (Right), the VocalNet-M2 Talker consists of a Talker backbone followed by \(N_{\text{mtp}}\) sequential MTP layers. This design allows the model to predict \(N_{\text{mtp}}+1\) audio tokens for each codebook in a single inference step while preserving the essential temporal relationships between these tokens.

Specifically, for an MTP Layer \(n\) (where \(n \in \{1, \ldots, N_{\text{mtp}}\}\)), it is designed to predict the audio tokens at time step \(t+n+1\), leveraging information available up to time \(t\). The output of an MTP layer can be formally expressed as:



\begin{equation}
\{a^{\text{cb}j}_{t+n+1}\}_{j=1}^{8} = \mathcal{T}_{\text{MTP}_n}\cdots \mathcal{T}_{\text{MTP}_1}\mathcal{T}_{\text{talker}}(h^{\text{up}}_{1:t} + \sum_{j=1}^{8}\mathrm{Emb}(a^{\text{cb}j}_{1:t}))
\end{equation}

As analyzed in VocalNet~\cite{wang2025vocalnet}, this approach helps the model to more efficiently leverage limited training data and better capture local dependencies between speech tokens. Consequently, the overall training objective integrates the standard Talker loss with losses from each MTP layer:


\begin{equation}
\begin{aligned}
    \mathcal{L}_{\text{mtp}} = -\sum_{n=0}^{N_{\text{mtp}}}\sum_{t=0}^{M-1}\sum_{j=1}^{8}\mathrm{log}P(a^{\text{cb}j}_{t+n+1}|h^{\text{up}}_{1:t}, \{a^{\text{cb}i}_{1:t}\}_{i=1}^{8})
\end{aligned}
\end{equation}

\subsection{Training Stategy}

The training strategy for VocalNet-M2 is structured in three sequential stages. The initial stage focuses on pre-training the VocalNet-M2 Talker using TTS data, where the model learns to synthesize high-quality speech solely from text tokens, thereby establishing its fundamental speech generation capabilities. Following this, the second stage is dedicated to training the downsample adaptor and VocalNet-M2 Thinker with Lora, enabling it to comprehend and process raw audio inputs and generate text responses. The final stage integrates both the Thinker and Talker modules for end-to-end fine-tuning on speech-to-speech dialogue data. Distinct from the initial TTS pre-training, here the Talker module receives both the hidden states and text embeddings generated by the Thinker. This comprehensive final phase allows the entire VocalNet-M2 model to process audio input, produce a pertinent textual response, and concurrently generate the corresponding speech output.


\section{Experimental Settings}
\label{sec:Experimental_Settings}

\subsection{Training Data}
\label{sec:taining_data}
For TTS pre-training, we used approximately 10k hours of randomly sampled audio from the Emilia dataset~\cite{he2024emilia}.
Our speech dialogue training dataset totals about 800K samples (approximately 7k hours of audio). This includes 400K dialogues from VoiceAssistant~\cite{xie2024mini}, 300K from Ultrachat~\cite{wang2025vocalnet}, and an additional 100K English multi-turn speech dialogues synthesized from tulu-3-sft-mixture~\cite{lambert2024tulu} using Cosyvoice2~\cite{du2024cosyvoice}.

Regarding model initialization, the audio encoder is based on the Whisper-large-v3~\cite{radford2023robust}. The VocalNet-M2 Thinker is initialized from Qwen3-8B~\cite{qwen3}. The VocalNet-M2 Talker shares a similar architectural design with the Thinker but employs a reduced number of transformer layers and  is trained separately from scratch. As for audio labels, we ultilize XY-tokenizer~\cite{gong2025xy} to extract tokens.

\subsection{Evaluation Metrics}
\label{sec: evaluation}
To thoroughly assess VocalNet-M2's voice interaction capabilities, we utilize the English subsets of OpenAudioBench~\cite{li2025baichuan}. The quality of textual responses generated by the model is evaluated using Qwen-max, which scores correctness and relevance on a normalized scale of 0 to 10. For evaluating speech quality, we employ two distinct metrics. UTMOS~\cite{saeki2022utmos} predicts Mean Opinion Scores (MOS) for objective measurement of perceived naturalness. To quantify the alignment between synthesized speech and its text, we transcribe the speech using Whisper-large-v3~\cite{radford2023robust} and then compute the Word Error Rate (WER) against the ground-truth text.

\section{Experimental Results}
\label{sec:Experimental_Results}
\subsection{Main Results}
Table~\ref{tab:performance_metrics} presents a comparative analysis between VocalNet-M2 and other mainstream SLMs. VocalNet-M2 demonstrates strong performance in text quality, retaining the knowledge and reasoning capabilities of Qwen3-8B. It achieves the highest scores in AlpacaEval and Llama Questions, alongside competitive results in TriviaQA and Web Questions. For speech quality, evaluated using UTMOS and WER, VocalNet-M2's performance is in the mid-range among current mainstream models, as shown in Table~\ref{tab:performance_metrics}. This is an expected outcome given the limited training data and the inherent complexity of learning multi-codebook speech tokens compared to single-codebook approaches.

Furthermore, we conducted a latency analysis, measuring the first audio chunk generation time for all models. To ensure a fair comparison and minimize variability arising from custom implementations, we prioritized models with officially provided streaming inference code. Consequently, Qwen2.5-Omni was excluded due to the absence of such code. Most models were evaluated using a fixed first chunk duration of 0.8 seconds; the sole exception was MiniCPM-o, whose official implementation has a fixed chunk size of 0.533 seconds. All tests were performed on a single L20 GPU without acceleration frameworks (e.g., vLLM), as these are not universally supported by the baseline models.
As detailed in Table~\ref{tab:performance_metrics}, VocalNet-M2 exhibits significantly lower first chunk latency while maintaining competitive performance. This efficiency is attributed to its direct modeling of multi-codebook speech tokens and the incorporation of the MTP approach.

\subsection{Comparison Between Single and Multi-Codebook Tokens}

This section explores the differences in modeling single and multi-codebook speech tokens. For the tokenizer with single-codebook, we utilize the $\mathcal{S}^3$ tokenizer from Cosyvoice2~\cite{du2024cosyvoice}. Conversely, the XY-tokenizer with 8 codebooks is utilized for extracting multi-codebook tokens. Two versions of training data were constructed:
\begin{itemize}
    \item \textbf{v1 data:} The speech dialogue data described in Section~\ref{sec:taining_data}.
    \item \textbf{v2 data:} A high-quality dialogue dataset of approximately 400K samples, derived from v1. It exhibits higher UTMOS and lower WER, achieved by filtering samples with high WER and re-synthesizing audio with high-quality prompts.
\end{itemize}
The results of this ablation study are presented in Table~\ref{tab:tokenizer_impact}.

Firstly, we observe that learning to generate multi-codebook speech tokens requires more training data than single-codebook tokens to achieve comparable performance. Without TTS pretraining, the model struggles to generate the multi-codebook tokens, resulting in very high WER and low UTMOS scores. In contrast, learning speech tokens extracted by the $\mathcal{S}^3$ tokenizer is easier; even without pretraining, the model can achieve respectable performance.

Secondly, high-quality training data is crucial for learning to generate multi-codebook speech tokens. Models trained with `emilia + v2' data demonstrate significantly better WER and UTMOS scores compared to those trained with `emilia + v1' data. This improvement is not observed in models leveraging single codebook speech tokens. The reason for this disparity is that single codebook speech tokens inherently lack acoustic information, which must be reconstructed through a flow-matching model to convert them back into speech. Consequently, the quality of speech generated from single codebook tokens heavily relies on the performance of the flow-matching model. In this work, we utilize the CosyVoice2 flow-matching model, enabling single-codebook-based models to achieve high performance in both WER and UTMOS, even when trained on noisier, lower-quality `v1' data. In contrast, models based on multi-codebook tokens directly learn acoustic features from the dialogue data, making them more sensitive to the quality of the training data.


\begin{table}[t]
\centering
\caption{Impact of tokenizer type (single vs. multi-codebook) and training data quality on speech generation performance.}
\resizebox{1\linewidth}{!}{
\begin{tabular}{llcccl}
\toprule
{\textbf{tokenizer}} & {\textbf{Training data}} & {\textbf{WER}} & {\textbf{utmos}} & {\textbf{First chunk latency}} \\
\hline
\multirow{3}{*}{$\mathcal{S}^3$  tokenizer(Single-codebook)} & v1 & 10.66 & 4.34 & \\
 & emilia + v1 & 3.73 & 4.35 & 725.90 $\pm$ 9.17 \\
 & emilia + v2 &  \textbf{3.68} & \textbf{4.37} & \\
\hline
\multirow{3}{*}{XY-tokenizer(Multi-codebook)} & v1 & 20.49 & 3.89 & \\
 & emilia + v1 & 10.43 & 4.08 & \textbf{405.23 $\pm$ 6.29} \\
 & emilia + v2 & 8.56 & 4.24 & \\
\bottomrule
\end{tabular}
}
\label{tab:tokenizer_impact}
\end{table}

\begin{table}[bt]
\centering
\caption{Ablation study on the impact of MTP on model's performance. }
\resizebox{1.0\linewidth}{!}{
\begin{tabular}{ccccccc}
\toprule
\multirow{2}{*}{\textbf{Metrics}} & \multirow{2}{*}{\textbf{w/o MTP Layer}} & \multicolumn{5}{c}{\textbf{w/ n MTP layer}} \\
\cmidrule(lr){3-7}
 & & n=1 & n=2 & n=3 & n=4 & n=5 \\
\midrule
WER &8.56 & 7.64 & 6.53 & 6.64 & 6.07 & 6.33\\
UTMOS & 4.24 & 4.28 & 4.29 & 4.28 & 4.31 & 4.29\\
\bottomrule
\end{tabular}
}
\label{tab:mtp_performance}
\end{table}

\subsection{Ablation study on MTP}
This section investigates the impact of MTP on the model's performance. In this experiment, we varied the number of MTP layers used during training, while MTP layers were not employed during inference. As shown in Table~\ref{tab:mtp_performance}, the introduction of MTP significantly reduced the WER from 8.56 (without MTP layers) to an optimal 6.07 with four MTP layers. UTMOS scores remained consistently high, with the best score of 4.31 also achieved with four MTP layers. Therefore, a configuration with four MTP layers was adopted for VocalNet-M2.

\subsection{Latency Analysis}
\begin{figure}[t]
\begin{minipage}[b]{1.0\linewidth}
  \centering
  \centerline{\includegraphics[width=1.0\textwidth]{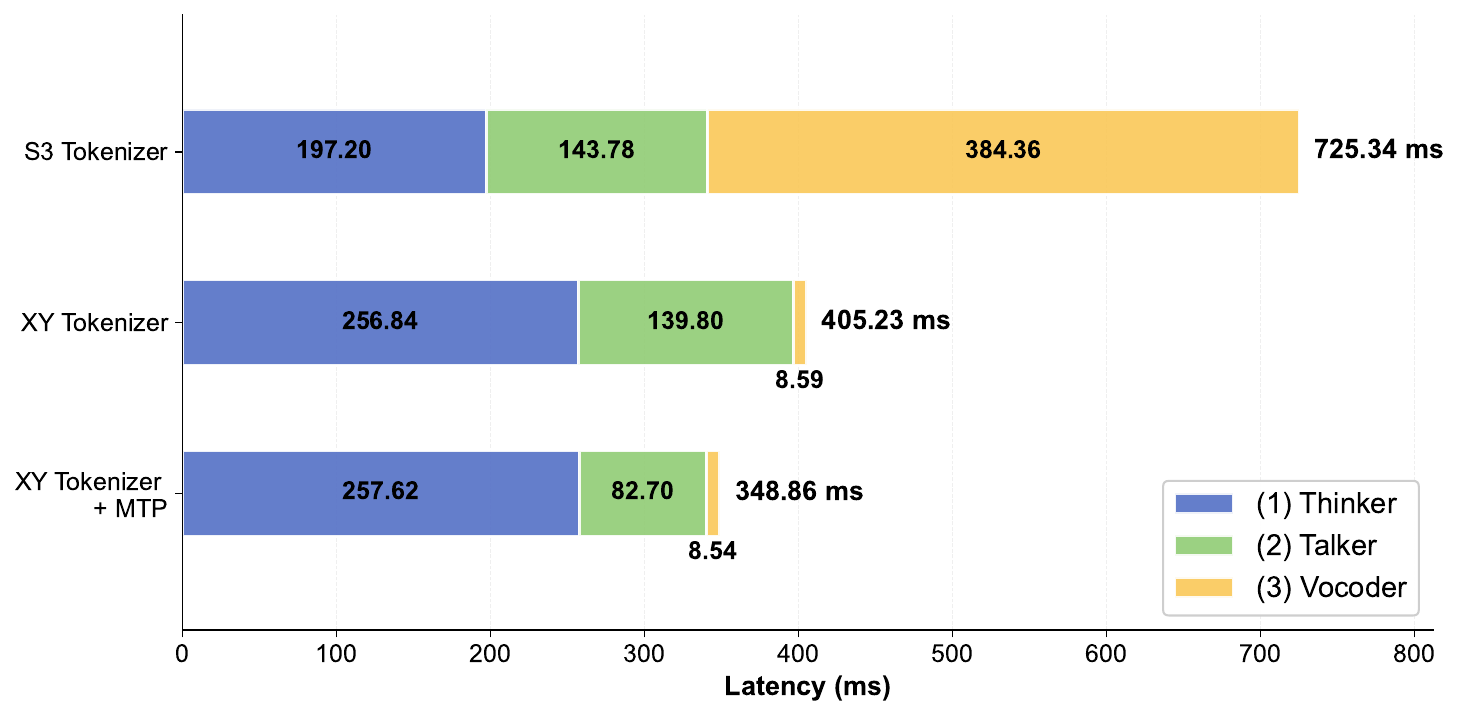}}
  \caption{Breakdown of first chunk latency across different model components for various configurations.}
  \label{fig:latency}
\end{minipage}
\end{figure}

We categorize the first chunk latency into three distinct parts: (1) the Thinker's generation of text tokens and hidden states, (2) the Talker's generation of speech tokens, and (3) the Vocoder's conversion of these tokens into speech. Figure~\ref{fig:latency} illustrates the latency for various configurations. Notably, the integration of multi-codebook speech tokens and the MTP method significantly reduced the latency in both the (2) Talker and (3) Vocoder stages. These advancements collectively enabled VocalNet-M2 to achieve a first chunk latency reduction from 725 ms to 348 ms, resulting in an inference speedup of approximately $2\times$.

\vspace{-0.2cm}

\section{Conclusion}
In this work, we introduced VocalNet-M2, a novel low-latency modality-alignment SLM. Our key contributions include a new model architecture designed to directly generate multi-codebook speech tokens, thereby eliminating the need for a computationally intensive flow-matching model for speech synthesis and significantly reducing latency. Additionally, we developed a MTP strategy that not only enhances overall performance but also further reduces inference latency. Our experimental results demonstrate that VocalNet-M2 effectively reduces first chunk latency by approximately 50\%, from 725ms to 348ms, showcasing a substantial improvement in responsiveness. We also provided a thorough comparison between single and multi-codebook approaches, highlighting their respective advantages and limitations. These advancements pave the way for more efficient and responsive spoken dialogue systems.


\bibliographystyle{IEEEbib}
\bibliography{strings,refs}

\end{document}